\documentclass[DTMColor]{ROB-New}

\usepackage{graphicx}
\usepackage{gensymb}
\usepackage{subfigure}
\usepackage{booktabs}
\usepackage{multirow}
\usepackage{xcolor}
\usepackage{amsmath}
\usepackage{mathrsfs}

\begin{document}

\authormark{Cambridge Authors}

\articletype{RESEARCH ARTICLE}

\jnlPage{1}{7}
\jyear{2021}
\jdoi{10.1017/xxxxx}

\title{FabricFolding: Learning Efficient Fabric Folding without Expert Demonstrations}

\author[1,2]{Can He}
\address[1]{Shenzhen Key Laboratory of Robotics Perception and Intelligence, Department of Electronic and Electrical Engineering, Southern University of Science and Technology, Shenzhen, China}
\address[2]{Jiaxing Research Institute, Southern University of Science and Technology, Jiaxing, China}

\author[1]{Lingxiao Meng}

\author[1,2]{Zhirui Sun}

\author[1,2]{Jiankun Wang}

\author[1]{Max Q.-H. Meng}

\address{\textbf{Corresponding authors:} Jiankun Wang, Max Q.-H. Meng;
\email{wangjk@sustech.edu.cn, max.meng@ieee.org}}

\received{xx xxx xxx}
\revised{xx xxx xxx}
\accepted{xx xxx xxx}


\abstract{Autonomous fabric manipulation is a challenging task due to complex dynamics and potential self-occlusion during fabric handling. 
An intuitive method of fabric folding manipulation first involves obtaining a smooth and unfolded fabric configuration before the folding process begins. 
However, the combination of quasi-static actions such as pick \& place and dynamic action like fling proves inadequate in effectively unfolding long-sleeved T-shirts with sleeves mostly tucked inside the garment. 
To address this limitation, this paper introduces an improved quasi-static action called pick \& drag, specifically designed to handle this type of fabric configuration. 
Additionally, an efficient dual-arm manipulation system is designed in this paper, which combines quasi-static (including pick \& place and pick \& drag) and dynamic fling actions to flexibly manipulate fabrics into unfolded and smooth configurations.
Subsequently, keypoints of the fabric are detected, enabling autonomous folding. 
To address the scarcity of publicly available keypoint detection datasets for real fabric, we gathered images of various fabric configurations and types in real scenes to create a comprehensive keypoint dataset for fabric folding. This dataset aims to enhance the success rate of keypoint detection.
Moreover, we evaluate the effectiveness of our proposed system in real-world settings, where it consistently and reliably unfolds and folds various types of fabrics, including challenging situations such as long-sleeved T-shirts with most parts of sleeves tucked inside the garment. 
Specifically, our method achieves a coverage rate of 0.822 and a success rate of 0.88 for long-sleeved T-shirts folding.}

\maketitle

\section{Introduction}

In recent years, significant progress has been made in the field of robotic manipulation, especially in the handling of rigid objects, and breakthroughs have been made in multiple aspects, such as the re-grasping of complex objects \cite{wu_fu_wang_2022, fukuda_2023} and manipulation in cluttered environments \cite{bejjani2021learning, ren2022rearrangement}. However, the autonomous manipulation of fabrics still faces significant challenges compared with rigid objects. This is mainly attributed to two key factors: the complex dynamic model of the fabric and the persistent self-occlusion problem during fabric manipulation. Therefore, further research is necessary to overcome these challenges and unlock the full potential of fabric manipulation for robotic applications.

Early fabric manipulation research aims to develop heuristic methods for quasi-static manipulations with a single robotic arm to accomplish tasks such as fabric unfolding \cite{triantafyllou2016geometric}, smoothing \cite{sun2014heuristic}, and folding \cite{doumanoglou2016folding}. 
However, these methods have some inherent limitations, including strong assumptions about the initial state and fabric type. 
Recently, there has been a surge in the development of deep learning techniques for fabric manipulation, where researchers introduce self-supervised learning to replace the need for expert demonstrations \cite{wu2020learning}. 
By learning goal-conditioned \cite{lee2021learning} strategies, it is now possible to effectively fold a single square fabric. 
However, this quasi-static method requires numerous iterations to obtain relatively smooth fabric. 
Ha \emph{et al.} \cite{ha2022flingbot} propose a dynamic fling method to achieve fabric unfolding, but it cannot be generalized to other tasks. 
Some researchers have attempted to combine supervised learning from expert demonstrations to realize fabric folding and unfolding \cite{speedfolding}, but these methods require extensive human annotations, which are time-consuming and error-prone.
Canberk \emph{et al.} \cite{clothfunnels} proposed a method that utilizes  canonicalized-alignment to unfold fabric, complemented by heuristic keypoint detection for folding the unfolded fabric. 
This approach solely relies on pick \& place and fling actions for unfolding the fabric. 
However, the proposed method encounters challenges in achieving effective unfolding when dealing with complex fabrics like long-sleeved T-shirts, particularly when certain sections of the sleeves are concealed or tucked beneath other fabric layers. Consequently, this limitation adversely impacts the fabric folding task. 
Moreover, due to the absence of publicly available keypoint datasets specifically designed for fabric folding, Canberk \emph{et al.} only collected 200 simulated cloth data samples for each fabric type to train the keypoint model. 
However, given the sim2real gap, employing simulation data for training purposes can result in inadequate detection accuracy and a lack of robustness in the detection model when applied to real-world scenarios. This limitation can significantly impact the effectiveness of subsequent cloth folding processes.

\begin{figure}[htbp]
    \centering
    \includegraphics[width= 0.9\columnwidth]{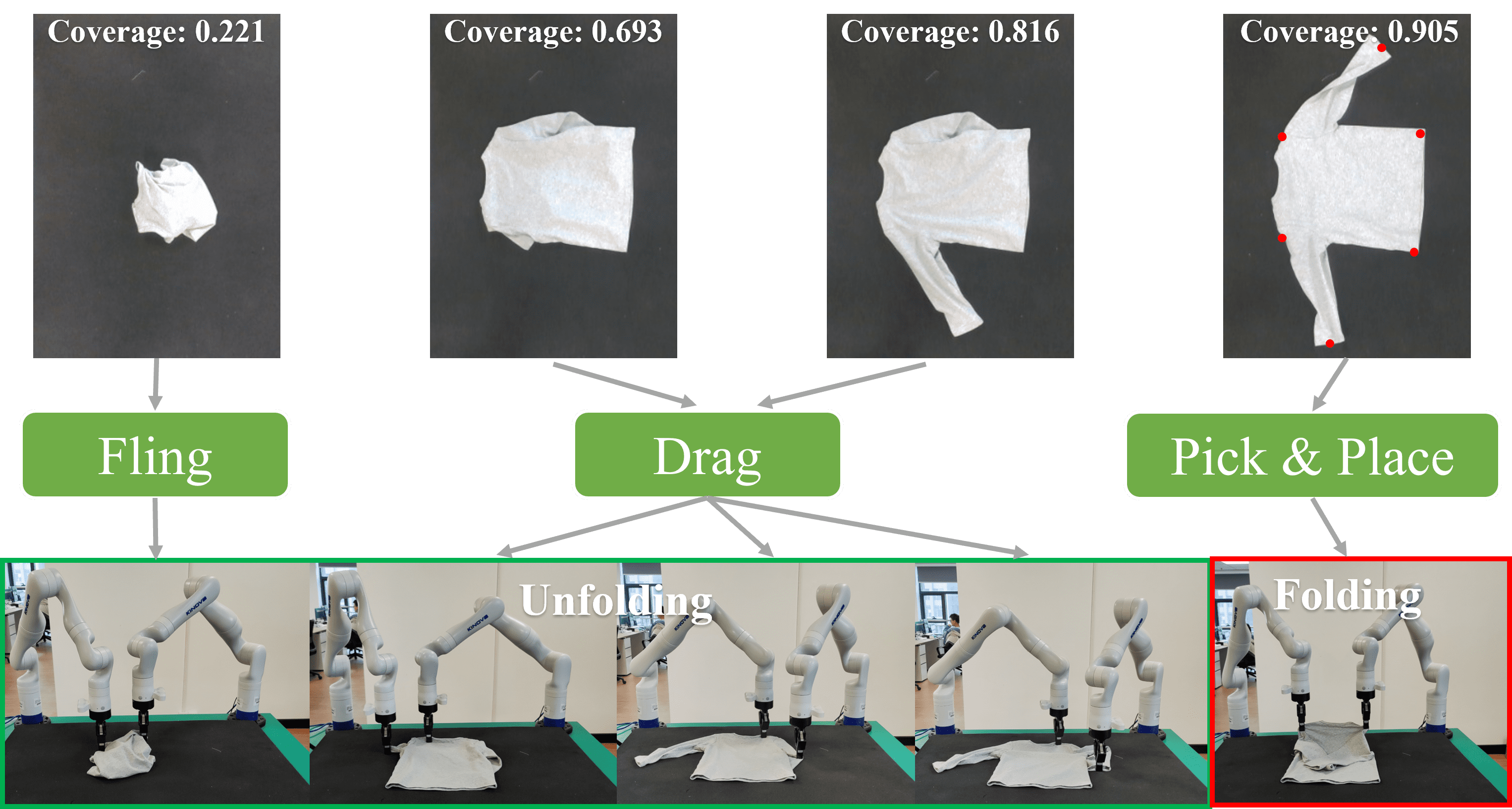}
    \caption{FabricFold divides the task of fabric folding into two stages for any given initial configuration. The first stage aims to achieve a relatively smooth fabric by dynamically selecting and executing actions (dynamic or quasi-static) based on the current state of the fabric. The second stage commences once the fabric is sufficiently unfolded, and involves performing the folding task by detecting the keypoints of the fabric}
    \label{example}
\end{figure}

This paper presents the introduction of a novel quasi-static motion primitive called pick \& drag. 
This primitive is designed to address situations where sections of long-sleeved T-shirts sleeves are concealed or tucked beneath other fabrics.  
Additionally, we propose FabricFolding, a system that leverages the fabric's current coverage and corresponding keypoint information to intelligently select suitable primitive actions for unfolding the fabric from arbitrary initial configurations and subsequently folding the unfolded fabric.
The system comprises two components: fabric unfolding and folding. 
The first component utilizes a self-supervised network to acquire knowledge of specific grasping points by analyzing RGBD input images. 
This knowledge is then applied to smoothen and unfold the fabric, which initially exists in a wrinkled state. 
The second component combines fabric keypoint detection with heuristic folding methods to enable efficient and precise folding of the fabric subsequent to its smoothing and unfolding.
Moreover, the design of our fabric keypoints network eliminates the need for
expert demonstrations and provides critical assistance during the fabric's unfolding stage.
Fig. \ref{example} shows the FabricFolding work example. 

The contributions of this paper are summarized as:

\begin{itemize} 
    \item We enhance a quasi-static primitive action, namely pick \& drag, to facilitate fabric unfolding, particularly when dealing with intricate cloth configurations like when sections of long-sleeved T-shirts are concealed or tucked beneath other layers of fabric.
    \item We propose a self-supervised learning unfolding strategy with a multi-manipulation policy that can choose between dynamic fling and quasi-static (including pick \& place and pick \& drag) actions to efficiently unfold the fabric.
    \item To bridge the sim2real gap and enhance the accuracy of fabric keypoint detection, we conduct an extensive data collection process involving real images of various fabric types. Subsequently, we curate a dedicated dataset specifically designed for fabric folding keypoint detection.
    \item Our method has been thoroughly evaluated using real robotic arms on a diverse range of fabrics. Specifically, it achieves a coverage rate of 0.822 and a fold success rate of 0.88 for long-sleeved T-shirts. Additionally, our method demonstrates a coverage rate of 0.958 and a fold success rate of 0.92 for towels.
\end{itemize}


\section{RELATED WORK}

\subsection{Fabric Unfolding}

Fabric unfolding mainly changes the fabric from an arbitrary crumpled configuration to a fully unfolded configuration. 
Prior work on fabric unfolding is based on heuristics and the extraction of some geometric features of the fabric, such as the edges \cite{triantafyllou2016geometric}, wrinkles \cite{sun2014heuristic, yuba2017unfolding}, and corners \cite{maitin2010cloth, seita2018robot} of the fabric. 
Then these features are utilized to determine the subsequent manipulation to make the fabric as smooth as possible. 
Recently, reinforcement learning combines hard-coded heuristics \cite{seita2020deep} or expert demonstrations \cite{ganapathi2021learning} to unfold fabric.
Wu \emph{et al.}  \cite{wu2020learning} introduce self-supervised learning into fabric unfolding to replace the role of expert demonstrations or heuristics. 
However, fabric unfolding  based on quasi-static manipulation may require numerous iterations and interactions before achieving a relatively smooth fabric. 
This method's effectiveness can be hampered by the limited reach of a robotic arm and a single gripper's operational constraints, rendering the task impractical, particularly for intricate fabrics like T-shirts. 

Compared with quasi-static manipulation, dynamic manipulation of robotic arms involves high-speed motion, which imparts velocity to the grasped deformable object. 
When the robotic arms come to a stop, the ungrasped portion of the deformable object continues moving due to inertia.
Dynamic manipulation effectively expands the reachable fabric area and reduces the number of operations necessary to accomplish the desired task.
Dynamic manipulation is initially applied to linearly deformable objects, such as cables \cite{lim2022real2sim2real}. Wang \emph{et al.} \cite{wang2020swingbot} employ tactile sensors to capture information about objects and integrate this data with recurrent neural networks to accomplish precise swinging movements. Jangir \emph{et al.} \cite{jangir2020dynamic} utilize reinforcement learning for dynamic operations on towels. However, this research was limited to simulations and did not include physical experiments.
Ha \emph{et al.} \cite{ha2022flingbot} propose a system based on self-supervised learning that dramatically increases the effectiveness of fabric unfolding by using high-speed fling action to smooth wrinkled fabric. 
Based on the research conducted by Ha \emph{et al.}, Xu \emph{et al.} \cite{xu2022dextairity} employ a commodity centrifugal air pump to smooth the fabric more efficiently.
Dynamic flings alone, however, cannot fully unfold complicated textiles, such as long-sleeved T-shirts. 
Avigal \emph{et al.} \cite{speedfolding} introduce the BiMaMa-Net architecture, which enables the selection of multiple primitive actions for fabric unfolding. 
Canberk \emph{et al.} \cite{clothfunnels} present a clothing alignment algorithm designed for cloth unfolding. This algorithm utilizes self-supervised learning to determine suitable primitive actions for cloth unfolding by encoding both the current fabric state and the aligned fabric state.
However, these algorithms generally do not account for intricate fabric configurations, such as when a portion of the long-sleeved T-shirt's sleeves is concealed beneath other layers of clothing.

\subsection{Fabric Folding}

Fabric folding initially relied on heuristic algorithms that impose strict predefined constraints on the fabric's initial configuration \cite{doumanoglou2016folding, cusumano2011bringing, garcia2022knowledge}. 
Contemporary approaches to fabric manipulation involve training goal-conditioned policies using reinforcement learning \cite{matas2018sim, yang2016repeatable, hietala2021closing, hietala2022learning}, self-supervised learning \cite{lee2021learning, huang2022mesh, hoque2022visuospatial}, and imitation learning \cite{seita2020deep} in either simulated \cite{mo2022foldsformer, thach2022learning, weng2022fabricflownet} or real robotic arms \cite{lee2021learning}. 
However, the strategy of employing simulation data for training essentially cannot achieve the desired effect of the simulation environment due to the sim2real gap on the robotic manipulator \cite{matas2018sim}. 
Moreover, a gap remains in the ability to generalize the approach to various types of fabrics \cite{weng2022fabricflownet}. 
To generalize fabric folding across different types of fabrics, some researchers \cite{clothfunnels, lips2022learning} proposed an approach that combines the detection of fabric keypoints with a heuristic for the folding process. 
However, owing to the absence of a dataset containing fabric keypoints derived from real cloth images, their study relies solely on simulated image data to generate diverse fabric keypoints datasets. 
This limitation affects the accuracy of fabric keypoints detection when applied in real-world scenarios.

\begin{figure*}[htbp]
\centering
\includegraphics[width = \columnwidth]{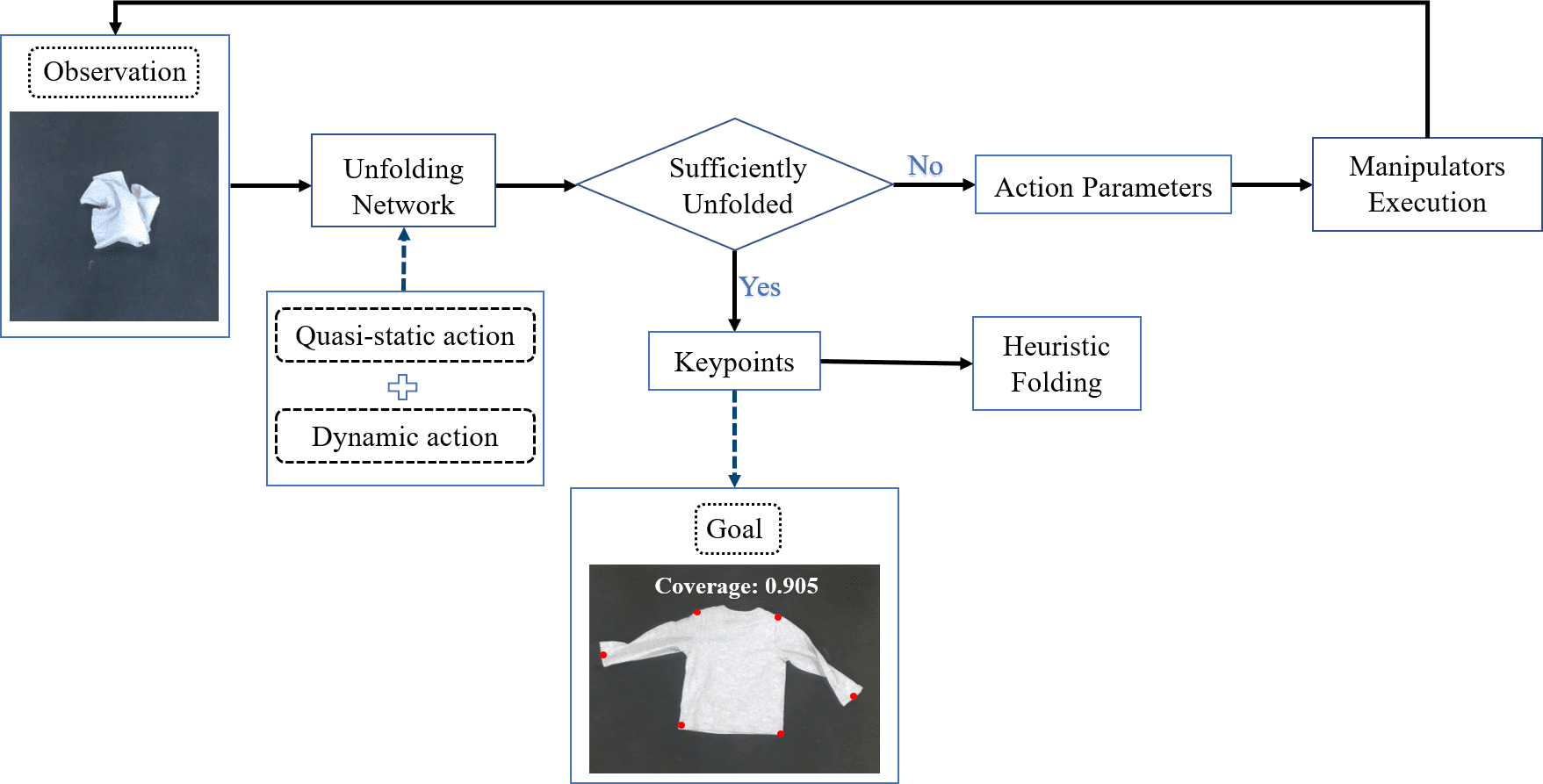}
\caption{\textbf{FabricFolding pipeline}: The RGB image and depth image obtained from an overhead camera serve as inputs to the unfolding network, which generates a set of keypoints and action parameters. If the fabric is sufficiently unfolded and corresponding keypoints are detected, the system will proceed to fold the fabric using heuristics based on the keypoints. Otherwise, the dual-arm system will execute relevant primitive actions to unfold the fabric}
\label{pipeline}
\end{figure*}

\section{METHOD}

FabricFolding is a novel approach for folding fabric, which involves breaking down the process into two distinct steps. 
Firstly, a multi-primitive policy unfolding algorithm is applied to ensure that the fabric is unfolded and smoothed out as much as possible. 
Subsequently, an adaptive folding manipulation technique is employed, which leverages keypoint detection to facilitate the folding of various types of fabric. 
FabricFolding can achieve effective and efficient fabric folding by utilizing this two-step approach, and the pipeline of the system is shown in Fig. \ref{pipeline}.

\subsection{Multi-primitive Policy Fabric Unfolding} 

While dynamic actions can efficiently unfold fabrics but cannot entirely unfold complex garments like long-sleeved T-shirt, quasi-static actions are preferable for delicately handling fabrics but are inefficient. 
To efficiently manipulate the fabric into a desired configuration, we utilize a combination of quasi-static and dynamic primitives that are capable of delicate handling and efficient unfolding, respectively.

FabricFolding utilizes the RGBD image output from an overhead RealSense 435i camera to calculate the current fabric coverage. The fabric coverage $\mathscr{C}$ is calculated as shown in Eq. (\ref{coverage}).

\begin{equation}
    \mathscr{C}=\frac{\sum_{i=1}^\mathcal{C} \mathcal{P}_i}{\sum_{j=1}^\mathcal{T}  \mathcal{P}_j}
    \label{coverage}
\end{equation}
where $\mathcal{P}_i$ represents the pixel occupied by the current fabric configuration in the image, while $\mathcal{P}_j$ represents the pixel occupied in the image when the fabric is fully unfolded.

\begin{figure*}[htbp]
\centering
\includegraphics[width = \columnwidth]{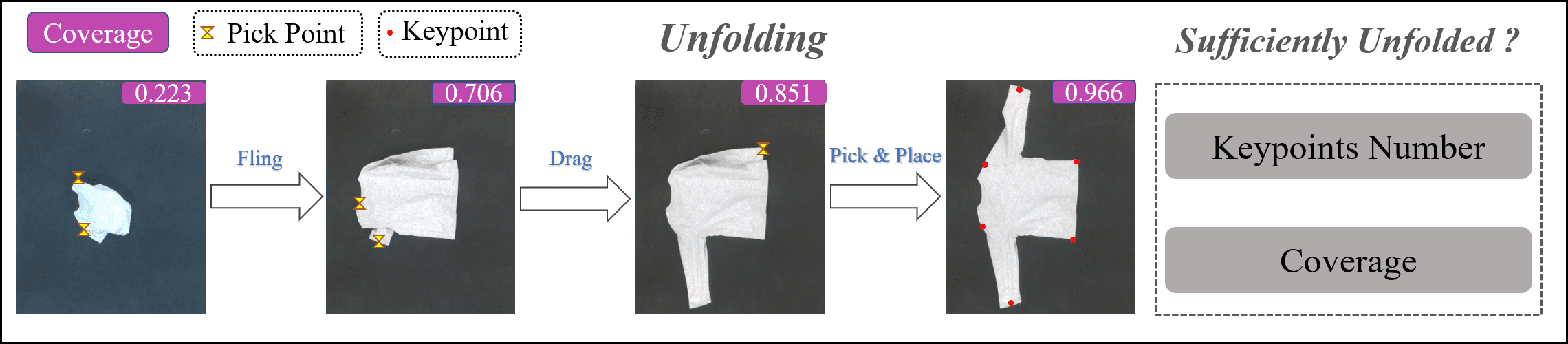}
\caption{\textbf{Fabric unfolding}: The system adopts a dynamic fling action to unfold the fabric when it is in a low-coverage stacking configuration. However, when the fabric coverage exceeds $S_1$, quasi-static actions such as pick \& place and pick \& drag are mainly used to fine adjustments. If the coverage of the fabric is greater than $S_2$ and the number of detected keypoints meets the requirements, the fabric is considered to be fully unfolded}
\label{unfolding}
\end{figure*}

To determine whether the fabric is ready for the downstream folding task, we have two indicators: the fabric coverage $\mathscr{C}$ and the number of detected keypoints $\mathcal{K}$. 
The fabric is considered smooth enough when the current coverage exceeds threshold $S_2$ and the number of detected keypoints meets the requirements. Fig. \ref{unfolding} illustrates the fabric unfolding pipeline with a multi-primitive policy. In our experiments, $S_1 = 0.65$ and $S_2 = 0.8$ perform best.

\subsubsection{Primitive Policy}

To determine suitable primitive actions for efficient fabric unfolding, We have devised a primitive action weighting strategy based on heuristics, and the precise formula is detailed in Eq. (\ref{policy}). 
When the coverage is below a certain threshold ($S_1$), the robotic arm will use the dynamic fling action with a high probability to effectively unfold the fabric. 
As the coverage of fabric increases and exceeds $S_1$, the system prefers to use quasi-static actions such as pick \& place and pick \& drag to operate the fabric. 
Finally, when the coverage is above ($S_2$), the robotic arms mainly rely on quasi-static actions to fine-tune the fabric. 
The action policy selection is shown below: 

\begin{equation}
\mathcal{A}(\mathscr{C}) = \begin{cases}0.99 a_d+0.01 a_{qs} &\!, \mathscr{C} \leqslant S_1 \\ \left(S_2-\mathscr{C}\right) a_d+\left[1-\left(S_2-\mathscr{C}\right)\right] a_{qs} &\!, S_1<\mathscr{C}<S_2 \\ 0.01 a_d+0.99 a_{qs} &\!, S_2 \leqslant \mathscr{C}\end{cases}
\label{policy}
\end{equation}
where $\mathcal{A}$ represents the chosen action, $a_{qs}$ means the quasi-static including two actions of pick \& place and pick \& drag, and $a_d$ means the dynamic primitive, fling. 

Furthermore, when choosing between the two quasi-static primitive actions, the system relies on the detection of keypoints on the fabric sleeves. 
Specifically, when the distance between the keypoint on the sleeve and the keypoint on the shoulder on the same side significantly deviates from the standard value, the pick \& drag primitive is chosen. 
Otherwise, the pick \& place primitive is selected. 
The precise formula is presented in Eq. (\ref{aqs}).
For example, For long-sleeved T-shirts where part of the sleeve is concealed or tucked under other fabric, the robotic arms perform the pick \& drag action to extract sleeves concealed beneath other fabrics, thereby increasing fabric coverage. Moreover, the pick \& place action is also a quasi-static manipulation used to sufficiently smooth the fabric.

\begin{equation}
a_{q s}= \begin{cases}a_{p d} & , \mathcal{K}_{s s} \&\left(d_{s s}<d_0 \right) \\ a_{p p} & , \text{others } \end{cases}
\label{aqs}
\end{equation}
where $a_{p d}$ means the pick \& drag action, $a_{p p}$ represents the pick \& place action, $\mathcal{K}_{ss}$ indicates whether the keypoints of the same side of the fabric are detected, such as the keypoints of the same side shoulder and sleeve of a long-sleeved T-shirt. Similarly, $d_{s s}$ represents the distance between two detected keypoints on the same side, and $d_0$ represents the distance between the corresponding two keypoints when the fabric is fully unfolded.

\begin{figure*}[htbp]
\centering
\includegraphics[width = \columnwidth]{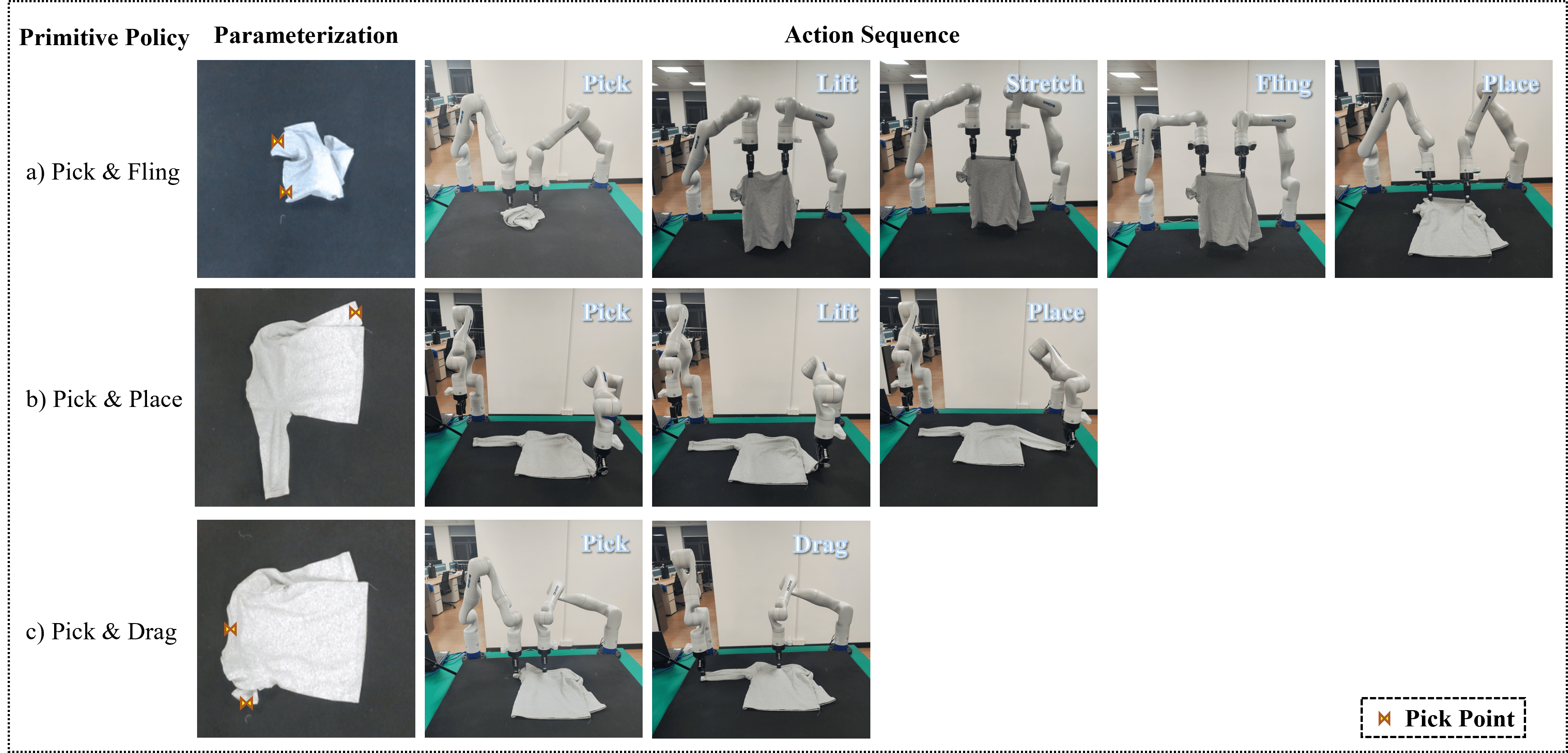}
\caption{\textbf{Primitive actions}: Based on the input received from the overhead RGBD camera, the grasping network can predict a series of pick poses for the upcoming primitive actions}
\label{primitve}
\end{figure*}

 All primitive actions are shown in Fig. \ref{primitve}, and the following are several primitive actions that we have defined:

\begin{itemize}
    \item \textbf{Pick \& Place: }
The pick \& place action is a quasi-static primitive. With a given pick pose and place pose, a single robotic arm grasps the fabric at the pick point, lifts it, moves it over the place point, and releases it. This primitive policy effectively handles situations where the hem or sleeves of the cloth are stacked on top of each other.
    \item \textbf{Pick \& Drag: }
It is a quasi-static primitive policy. Two pick poses are given, and the two robotic arms grasp the two pick points of the fabric, respectively. Then, one robotic arm remains stationary (close to the center of the fabric mask), while the other robotic arm drags the fabric away from the center point of the fabric mask for a certain distance. This primitive policy is effective in dealing with situations where most of the sleeves are concealed or tucked beneath other layers of cloth.
    \item \textbf{Fling: }
This is a dynamic primitive policy designed for fabric manipulation. Given two pick poses. After the robotic arms grasp the two pick points of the fabric, the fabric is lifted to a certain height and stretched, while the camera in front of the robotic arms is used to estimate whether the fabric has been fully stretched. Then the two robotic arms simultaneously fling the fabric forward for a certain distance, then retreat for a certain distance while gradually reducing the height, and then release the fabric. This policy efficiently spreads the fabric and increases coverage, but it may not be effective in dealing with smaller wrinkles in the cloth.
    \item \textbf{Fold: }
Both robotic arms execute the pick \& place primitive action simultaneously. The two pick poses and their corresponding place poses are obtained through keypoint detection of the fabric.
    
\end{itemize}

\subsubsection{Grasping Policy}

To enhance the effectiveness of fabric unfolding and ensure that the fabric becomes smoother, we have made enhancements to the grasping framework of DextAIRity \cite{xu2022dextairity}. These improvements include modifying the grasping action parameters and incorporating a primitive action selection block. These modifications allow for more accurate and efficient predictions of the grasp poses required for subsequent primitive actions.

    \begin{figure*}[htbp]
        \centering
        \includegraphics[width = 0.3\columnwidth]{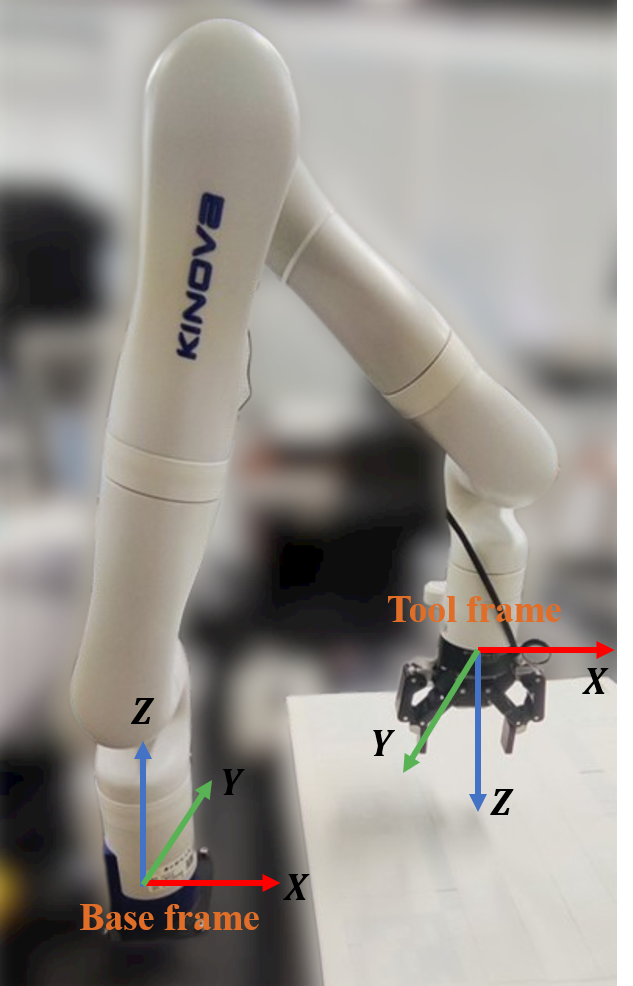}
        \caption{The left arm coordinate system in the two-arm system}
        \label{robotiq}
    \end{figure*}
    
\begin{figure*}[htbp]
\centering
\includegraphics[width = \columnwidth]{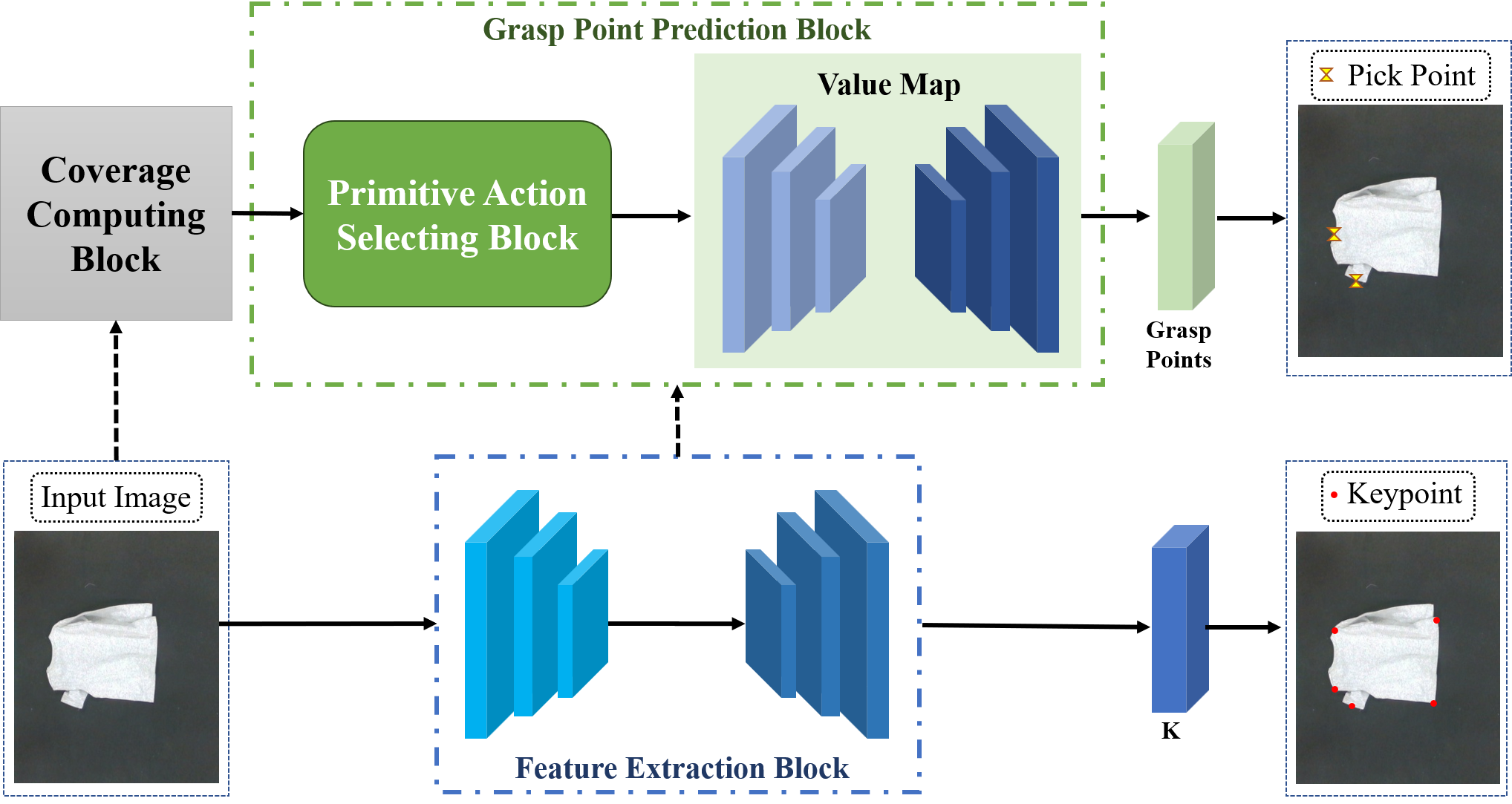}
\caption{\textbf{FabricFolding structure}: The RGB image captured by the overhead RealSense 435i camera serves as the input for our system}
\label{network}
\end{figure*}

\begin{itemize}
    \item \textbf{Grasping Action Parameterization: }
    DextAIRity \cite{xu2022dextairity} made some adjustments to Flingbot's \cite{ha2022flingbot} action parameterization, extending the two grasping positions L and R to the edge of the fabric mask, thereby reducing the likelihood of the gripper grasping multiple layers of fabric. 
    Based on the action parameterization form of DextAIRity ($C$, $\theta$, $\omega$), we have made some minor adjustments, and our action parameters are shown in Eq. (\ref{action parameterization}).
    \begin{equation}
        \mathcal{P}_{epo} = (C,\theta,\omega, \phi)
        \label{action parameterization}
    \end{equation}
     Where The meaning of ($C$, $\theta$, $\omega$) corresponds to its definition in DextAIRity, $\phi$ represents the angle of rotation of the manipulator's end effector relative to the y-axis of the base frame. For a visual representation of the robot arm end effector's coordinate system, please refer to Fig. \ref{robotiq}. This parameter primarily serves the purpose of adjusting the clamping claw's angle to facilitate the robot arm in grasping the fabric more effectively.

    \item \textbf{Unfolding Network: }
    We have designed an image feature extraction block based on Unet \cite{unet}, which supplies both predicted fabric keypoints and other image feature data to the grasp point prediction block.
    This grasping point prediction block comprises two integral parts: the primitive action selection block and the value map block. Specifically, to enhance the efficiency of fabric unfolding, the primitive action selection block assesses and selects appropriate primitive actions based on the current fabric coverage and the detected keypoints.
    Once the primitive action is determined, a set of action parameterizations denoted as $\mathcal{P}_{epo}$ is acquired through the action value map module, where the parameters with the highest values serve as the grasping parameters. To achieve equivariance between the grasping action and the physical transformation of the fabric, we employ a spatial action map \cite{ha2022flingbot, wu2020spatial}.
    The network structure is shown in Fig. \ref{network}.
        
\end{itemize}

\subsubsection{Training Setting}

Before training the grasping network, the keypoint detection network is trained first through supervised learning. 
We employed a simulation environment based on SoftGym \cite{lin2021softgym} for the pre-training of the fabric-unfolding grasping network through self-supervised learning. This primarily involved training three parameters ($C$, $\theta$, $\omega$). Due to the simulation environment's inability to import the URDF models of the robotic arms, we utilized two grasp points to represent the end effectors of the two robotic arms. Certain constraints were imposed on these points to ensure their applicability to the physical robotic arms. 
Subsequently, the network parameters acquired during simulation training were further trained in real-world conditions. Parameter $\phi$ was introduced during this stage to facilitate the fabric gripping by the robot arm. To prevent potential damage to the cameras of the two-arm system during manipulation, we imposed a constraint requiring the left and right robotic arms to rotate 90° and -90°, respectively, relative to the base frame's Z-axis. This ensured that the two cameras faced in opposite directions.

\begin{figure}[htbp]
\centering
\subfigure[Towel]{\includegraphics[width=.28\columnwidth] 
{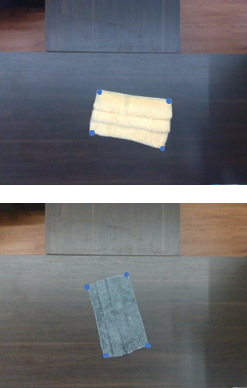} \includegraphics[width=.28\columnwidth] 
{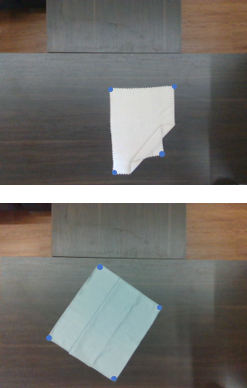}}
\subfigure[Long-leeved T-shirt]{\includegraphics[width=.28\columnwidth]{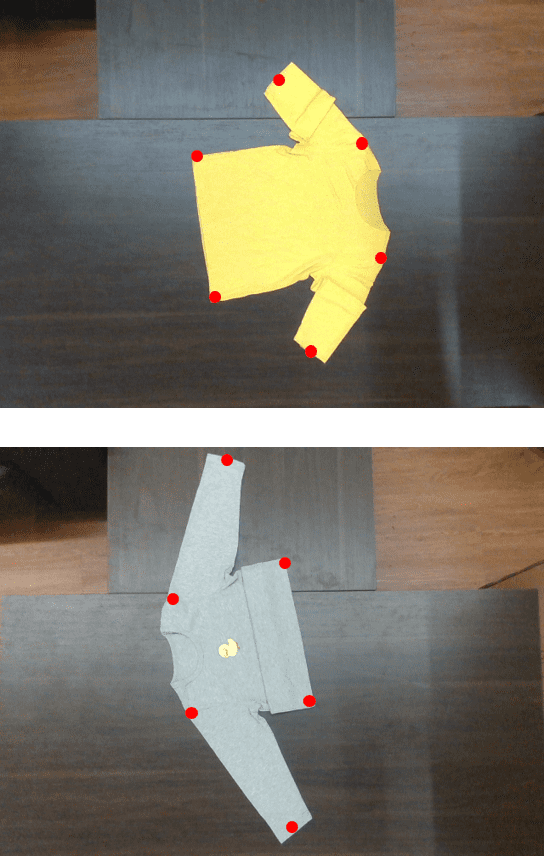}}
\caption{Sample keypoint detection dataset for fabric folding}
\label{dataset}
\end{figure}

\subsection{Heuristic Fabric Folding}

\subsubsection{Dataset for Keypoint Detection}

Due to the sim2real gap, the keypoint detection performance of the fabric is not optimal when using the fabric keypoint dataset generated in the simulation, which includes various configurations of fabric. For example, certain configurations of fabric cannot detect all the expected keypoints. Additionally, there is no existing public dataset that is suitable for keypoint detection in fabric folding tasks. In order to address this issue, we created a fabric keypoint dataset consisting of 1809 images that include four types of long-sleeved T-shirts and ten types of towels. Among them, all fabrics are configured to have a coverage rate of more than 65$\%$. For each fabric configuration, the long-sleeved T-shirt is rotated $360\degree$ for sampling, with each rotation being 10\degree; while the towel is rotated 180\degree, with each rotation being 18\degree.  

For towels, we define four keypoints, starting with the upper left corner of the image as corner1, and continuing in clockwise order as corner2, corner3, and corner4. Long-sleeved T-shirts have five key points, including right\_shoulder, left\_shoulder, right\_sleeve, right\_waist, left\_waist, and left\_sleeve, which is marked according to the outward direction of the vertical image. An example of the fabric keypoint dataset can be seen in Fig. \ref{dataset}.

\subsubsection{Keypoint Detection Network}

The keypoint detection network we designed uses Unet \cite{unet} as the backbone. We divide the data into training and validation sets in an 8:2 ratio. After training for 4 hours on an NVIDIA RTX3080Ti, the average pixel error of the detected keypoints on a 640x480 validation set image can be guaranteed to be within 3 pixels.

\subsubsection{Heuristic Folding}
\begin{figure*}[htbp]
\centering
\includegraphics[width = \columnwidth]{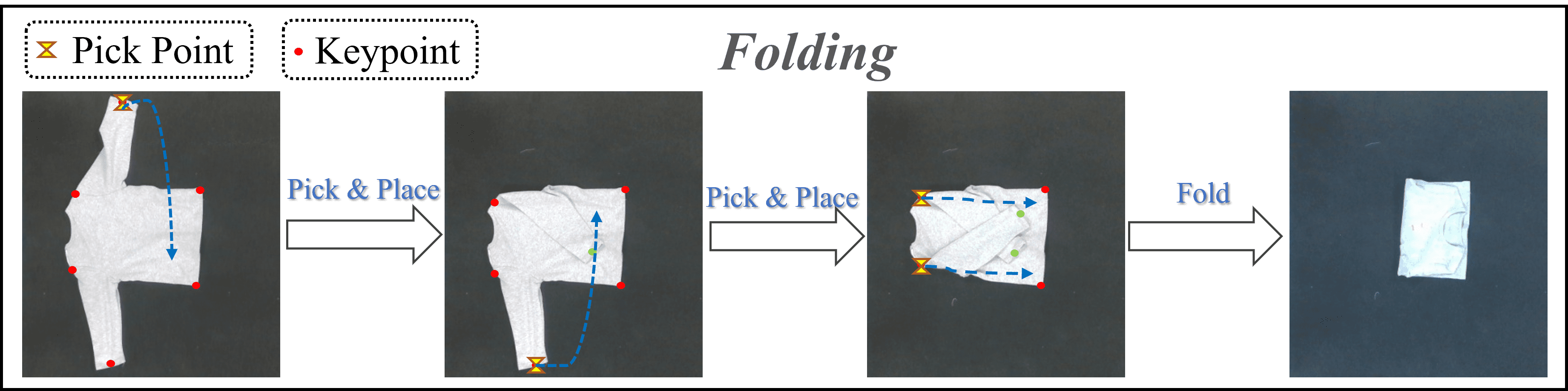}
\caption{\textbf{Fabric Folding}: Based on the input received from the overhead RGBD camera, the grasp network can predict a series of pick poses for the upcoming primitive actions}
\label{fold}
\end{figure*}

Taking inspiration from the Cloth Funnels \cite{clothfunnels} heuristic folding method, as depicted in Fig. \ref{fold}, we use a similar approach for folding a long-sleeved T-shirt. Initially, we utilize the pick \& place primitive action to fold the two sleeves of the garment onto the main part of the garment. Following this, we pick the keypoints of the shoulders and place them at the keypoints of the waist.

\section{EVALUATION}

\subsection{Metrics}

We evaluate FabricFolding on both the fabric unfolding and fabric folding tasks. Performance in the fabric unfolding task is evaluated based on the coverage achieved at the end of each episode. Furthermore, we evaluate the algorithm's success rate for the folding task, as well as its generalization ability to handle unseen fabrics on real robotic arms. To reduce experimental randomness, each experiment is repeated 25 times, and the weighted average of all results is calculated. If the folding result is deemed a failure by the majority of the 5 judges or if the folding cannot be completed after 20 action sequences, the experiment is considered a failure.

\subsection{Fabric Unfolding}
\begin{figure}[htbp]
\centering
\includegraphics[width = 0.8\columnwidth]{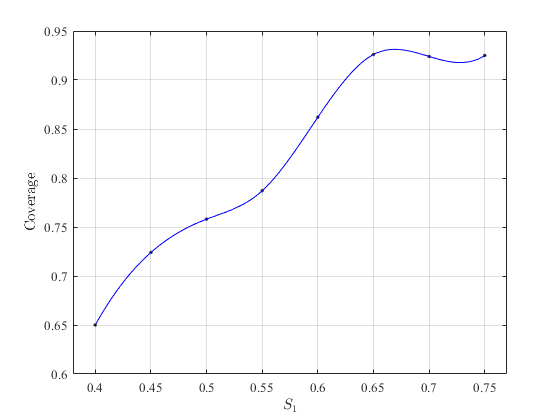}
\caption{The normalized coverage of a long-sleeved T-shirt (Initial coverage is 0.5) is evaluated after 5 primitive actions under various threshold parameters $S_1$}
\label{s1}
\end{figure}

To optimize the grasping network, we use $S_1$ as a threshold to select between dynamic and quasi-static primitives. To determine the optimal value for $S_1$, we compare the coverage achieved after 5 primitive actions for different values of $S_1$. As shown in Fig. \ref{s1}, the best folding efficiency is achieved when $S_1 = 0.65$.

Dynamic action can efficiently unfold the fabric, and quasi-static actions can make some fine adjustments to the fabric. To verify the effectiveness of our multi-primitive policy mechanism, we conduct experiments on long-sleeved T-shirts with different coverage. Table \ref{effective} shows that when the fabric has high initial coverage and mild self-occlusion, the quasi-static actions have better coverage compared to the dynamic action. On the other hand, when the fabric has low initial coverage and severe self-occlusion, the dynamic action effectively improves coverage. This observation is in line with the findings of Flingbot \cite{ha2022flingbot}. It also demonstrates that our multi-primitive policy outperforms a single primitive action in achieving higher coverage, regardless of the initial coverage of the fabric. This highlights the effectiveness of the multi-primitive policy in enhancing fabric coverage.

\begin{table}[htbp]
\centering
\caption{Validity of primitive actions: the fabric is a long-sleeved T-shirt}
\begin{tabular}{@{}ccc|l@{}}
\toprule
\textbf{Fabric Status}                        & \multicolumn{2}{c|}{\textbf{Primitive Actions}}                  & \textbf{Cov. $\uparrow$}  \\ \midrule
\multirow{5}{*}{Ini\_cov $\geq 0.7$} & \multirow{3}{*}{Quasi-static primitive action} & Pick \& Place(P\&P)   & 0.872          \\
                                       &                                         & Pick \& Drag(P\&D)    & 0.876          \\
                                        &                                         & P\&P + P\&D   & 0.891 \\
                                       & Dynamic primitive action                      & Fling           & 0.818          \\
                                       & Muti-primitive actions                          & P\&P+P\&D+Fling & \textbf{0.902} \\\midrule
\multirow{5}{*}{Ini\_cov $\leq 0.3$} & \multirow{3}{*}{Quasi-static primitive action} & Pick \& Place(P\&P)   & 0.624          \\
                                       &                                         & Pick \& Drag(P\&D)    & 0.636          \\
                                       &                                         & P\&P + P\&D   & 0.694 \\
                                       & Dynamic primitive action                      & Fling           & 0.742          \\
                                       & Multi-primitive actions                         & P\&P+P\&D+Fling & \textbf{0.822} \\ \bottomrule
\end{tabular}
\label{effective}
\end{table}

To assess the validity of the enhanced Pick \& Drag primitive action in unfolding severely self-occluding fabrics, we conducted tests using a long-sleeved T-shirt, with certain sleeves concealed or tucked beneath other layers of cloth. As indicated in Table \ref{pd validation}, the enhanced pick \& drag primitive action outperforms the individual pick \& place or fling in efficiently unfolding this long-sleeved fabric.

\begin{table}[htbp]
\centering
\caption{Validity of Pick \& Drag: the fabric is long-sleeved T-shirts with sleeves partially concealed beneath other layers of cloth}
\begin{tabular}{@{}ccc|l@{}}
\toprule
\textbf{Fabric Status}                        & \multicolumn{2}{c|}{\textbf{Primitive Actions}}                  & \textbf{Cov. $\uparrow$}  \\ \midrule
\multirow{5}{*}{Ini\_cov $\geq 0.75$} & \multirow{3}{*}{Quasi-static primitive action} & Pick \& Place(P\&P)   & 0.816          \\
                                       &                                         & Pick \& Drag(P\&D)    & 0.879          \\
                                        &                                         & P\&P + P\&D   & 0.895 \\
                                       & Dynamic primitive action                      & Fling           & 0.811          \\
                                       & Muti-primitive actions                          & P\&P+P\&D+Fling & \textbf{0.899} \\ \bottomrule
\end{tabular}
\label{pd validation}
\end{table}

Table \ref{coveragetable} presents the coverage attained by various algorithms on diverse real-world fabrics with initial coverage less than $30\%$.
Unfortunately, despite the availability of open-source code for SpeedFold \cite{speedfolding}  and Canberk \cite{clothfunnels}, substantial challenges exist when it comes to deploying these two tasks in a dual-arm system using Kinova Gen3 due to inconsistencies in the robotic arms' utilization.
Therefore, the data presented in this table are the original results reported in their respective papers. 
To ensure a fair comparison, the fabrics tested in our algorithm are chosen to be as similar as possible to those used in the previous works. 
The results in Table \ref{coveragetable} imply that our algorithm outperforms the other algorithms in terms of unfolded coverage on both towels and long-sleeved T-shirts.

\begin{table}[htbp]
\centering
\caption{Real-world coverage: $*$ indicates the original data in paper}
\label{coveragetable}
\begin{tabular}{@{}ccc@{}}
\toprule
\textbf{Approach}          & \textbf{Fabric}                            & \textbf{Cov. $\uparrow$}    \\ \midrule
\multirow{2}{*}{Flingbot \cite{ha2022flingbot}}  & \multicolumn{1}{c|}{Towel}                 & 0.905             \\
                           & \multicolumn{1}{c|}{long-sleeved T-shirts} & 0.742             \\ \midrule
\multirow{2}{*}{SpeedFold \cite{speedfolding}} & \multicolumn{1}{c|}{Towel}                 & $0.92^*$             \\
                           & \multicolumn{1}{c|}{T-shirts} & $0.8^*$ \\ 
                           & \multicolumn{1}{c|}{long-sleeved T-shirts} & \textbackslash{}  \\ \midrule
\multirow{2}{*}{Canberk \cite{clothfunnels}}   & \multicolumn{1}{c|}{Towel}                 & \textbackslash{} \\
                           & \multicolumn{1}{c|}{long-sleeved T-shirts} & $0.806^*$            \\ \midrule
\multirow{2}{*}{Our}       & \multicolumn{1}{c|}{Towel}                 & \textbf{0.958}    \\
                           & \multicolumn{1}{c|}{long-sleeved T-shirts} & \textbf{0.822}   \\ \bottomrule
\end{tabular}
\end{table}

\begin{table}[htbp]
\centering
\caption{Folding's success: $*$ indicates the original data in paper}
\label{success}
\begin{tabular}{@{}ccc@{}}
\toprule
\textbf{Approach}          & \textbf{Fabric}                            & \textbf{Success $\uparrow$} \\ \midrule
\multirow{2}{*}{Doumanoglou \cite{doumanoglou2016folding}}  & \multicolumn{1}{c|}{Towel}                 & $0.78^*$             \\
                           & \multicolumn{1}{c|}{T-shirts} & $0.66^*$             \\ \midrule
\multirow{2}{*}{SpeedFold \cite{speedfolding}} & \multicolumn{1}{c|}{T-shirt}               & $\textbf{0.93}^*$    \\
                           & \multicolumn{1}{c|}{long-sleeved T-shirts} & \textbackslash{} \\ \midrule
\multirow{2}{*}{Canberk \cite{clothfunnels}}   & \multicolumn{1}{c|}{Towel}                 & \textbackslash{} \\
                           & \multicolumn{1}{c|}{long-sleeved T-shirts} & $0.878^*$            \\ \midrule
\multirow{2}{*}{Our}       & \multicolumn{1}{c|}{Towel}                 & \textbf{0.92}    \\
                           & \multicolumn{1}{c|}{long-sleeved T-shirts} & \textbf{0.88}    \\ \bottomrule
\end{tabular}
\end{table}

The success rate of fabric folding is a crucial performance indicator. As shown in Table \ref{success}, the complexity of the fabric has a direct effect on the success rate of folding, which decreases continuously as the fabric complexity increases. In particular, when dealing with fabrics that are not whole pieces, such as long-sleeved T-shirts with two sleeves, it is common for the fabric to self-occlude during the folding process to reduce the success rate. Among the four algorithms compared, our algorithm outperformed the others in terms of the success rate achieved in folding towels and long-sleeved T-shirts.

\section{CONCLUSIONS}

In this paper, We enhance a quasi-static primitive action, pick \& drag, enabling it to address severely self-occluding fabric scenarios, including instances such as long-sleeved T-shirts with sleeves concealed or tucked beneath other layers of cloth. Simultaneously, we have developed the FabricFolding system, which dynamically selects multiple primitive actions to efficiently unfold fabric in any initial configuration. Additionally, we have developed a keypoint detection dataset for fabric folding to enhance the precision of fabric keypoint detection, consisting of approximately 2,000 images. Our algorithm achieves a coverage rate of 0.822 and a folding success rate of 0.88 for long-sleeved T-shirts. During our experiments, we find that when the two pick points of fling are at diagonally opposite corners of the fabric, it can be challenging to fully unfold the fabric even after multiple interactions. In the future, we plan to investigate this issue and develop solutions to address it.

\bibliographystyle{roblike}
\bibliography{refs}

\end{document}